\pdfoutput=1

\documentclass[11pt]{article}

\usepackage[final]{acl}

\usepackage{times}
\usepackage{latexsym}

\usepackage[T1]{fontenc}

\usepackage[utf8]{inputenc}

\usepackage{microtype}

\usepackage{inconsolata}

\usepackage{graphicx}

\usepackage{enumitem}
\usepackage{booktabs}
\usepackage{multirow}

\title{CMU’s IWSLT 2025 Simultaneous Speech Translation System}

\author{
 \textbf{Siqi Ouyang}\quad
 \textbf{Xi Xu}\quad
 \textbf{Lei Li}
\\
 Language Technologies Institute, Carnegie Mellon University, USA
\\
 \texttt{\{siqiouya,xixu\}@andrew.cmu.edu}
}

\begin{document}
\maketitle

\begin{abstract}
This paper presents CMU's submission to the IWSLT 2025 Simultaneous Speech Translation (SST) task for translating unsegmented English speech into Chinese and German text in a streaming manner. Our end-to-end speech-to-text system integrates a chunkwise causal Wav2Vec 2.0 speech encoder, an adapter, and the Qwen2.5-7B-Instruct as the decoder. We use a two-stage simultaneous training procedure on robust speech segments curated from LibriSpeech, CommonVoice, and VoxPopuli datasets, utilizing standard cross-entropy loss. Our model supports adjustable latency through a configurable latency multiplier. Experimental results demonstrate that our system achieves 44.3 BLEU for English-to-Chinese and 25.1 BLEU for English-to-German translations on the ACL60/60 development set, with computation-aware latencies of 2.7 seconds and 2.3 seconds, and theoretical latencies of 2.2 and 1.7 seconds, respectively.

\end{abstract}
\section{Introduction}

CMU's submission to the IWSLT 2025 Simultaneous Speech-to-Text Translation track~\cite{abdulmumin-etal-2025-findings}\footnote{\url{https://iwslt.org/2025/simultaneous}} is an end-to-end model that effectively translates unbounded English speech input into German and Chinese text without speech segmentation.

Translating unbounded speech presents unique challenges. Unlike segmented speech translation, it requires the model to maintain and process the speech and translation history so that translation quality, theoretical latency, and computation cost can be balanced~\cite{papi-etal-2024-streamatt,10.1162/tacl_a_00740}. Large language models (LLMs) have recently shown strong performance in improving speech translation quality~\cite{zhang2023tuninglargelanguagemodel,10447553,huang2023speechtranslationlargelanguage,xu-etal-2024-cmus,ahmad-etal-2024-findings}, and modern LLMs now support long-context inference due to architectural and algorithmic advances~\cite{han-etal-2024-lm,rope}. These two advantages were recently unified in InfiniSST~\cite{ouyang2025infinisstsimultaneoustranslationunbounded}, which frames simultaneous translation as a multiturn dialogue and enables inference on unbounded speech with minimal computational overhead.

Our system is built upon the InfiniSST framework and consists of: 
\begin{enumerate}[align=left] 
\item A chunkwise causal Wav2Vec 2.0 Large encoder~\cite{NEURIPS2020_92d1e1eb}, which incrementally processes the unbounded speech input. 
\item The Qwen2.5-7B-Instruct LLM~\cite{qwen2025qwen25technicalreport}, which receives the encoded speech features and performs simultaneous translation using a specially designed key-value (KV) cache management strategy.
\end{enumerate}
However, a major limitation in speech translation research is the scarcity of high-quality parallel speech-text data. Only several hundred hours are available on resources such as EuroParl-ST~\cite{9054626} and CoVoST2~\cite{wang21s_interspeech}. To scale InfiniSST training beyond this constraint, we synthesize training data by translating transcripts from automatic speech recognition (ASR) datasets into target-language text using an LLM. 

Our experiments on the ACL60/60 development set~\cite{salesky-etal-2023-evaluating} demonstrate that increasing the amount of synthesized speech translation data consistently improves translation quality, with gains observed even beyond 3,000 hours of training data. Additionally, we find that Qwen2.5-7B-Instruct significantly outperforms Llama3.1-8B-Instruct~\cite{grattafiori2024llama3herdmodels} on English-to-Chinese translation and achieves comparable performance on English-to-German translation.

\begin{figure*}[t]
    \centering
    \includegraphics[width=\linewidth]{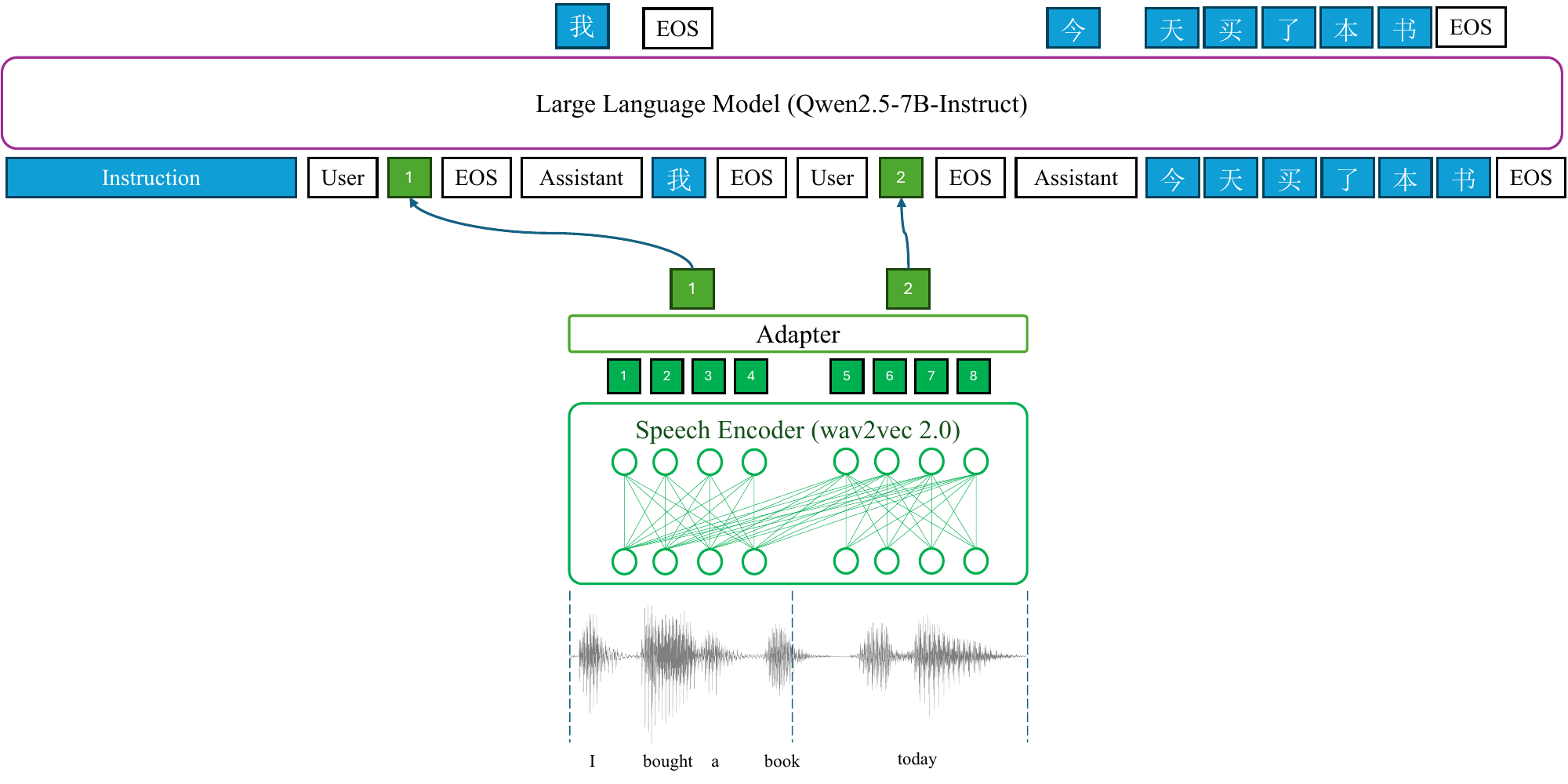}
    \caption{Model Architecture}
    \label{fig:InfiniSST-model}
\end{figure*}

\section{Task Description}

The IWSLT 2025 Simultaneous Speech-to-Text Translation track\footnote{\url{https://iwslt.org/2025/simultaneous}} focuses on translating unsegmented speech into target-language text using pretrained large language models (LLMs) and speech encoders. The evaluation data consists of unsegmented ACL talks. For English-to-German, systems are additionally tested on accented speech, while a dedicated development set is provided for Czech-to-English.

Systems are evaluated on both translation quality and latency. Latency is measured using StreamLAAL~\cite{papi-etal-2024-streamatt}, while translation quality is assessed using BLEU~\cite{papineni-etal-2002-bleu} and neural metrics such as COMET~\cite{guerreiro-etal-2024-xcomet}, BLEURT~\cite{sellam-etal-2020-bleurt}, etc.

We participate in two language directions: English-to-Chinese and English-to-German. For both directions, we submit models operating in the low-latency regime—achieving StreamLAAL $\leq$ 2 seconds for German and $\leq$ 2.5 seconds for Chinese on the development set.
\section{System Description}

\subsection{Model Architecture}

Our model architecture builds upon InfiniSST~\cite{ouyang2025infinisstsimultaneoustranslationunbounded}, a simultaneous speech translation system designed to efficiently handle unbounded streaming speech input and generate target text incrementally.
The architecture comprises three primary components: 1) a streaming speech encoder that incrementally computes representations from partial speech without redundant computations; 2) a speech-to-token embedding adapter that aligns speech representations with the LLM’s token embedding space; and 3) a multi-turn LLM decoder that dynamically processes speech inputs and produces translations interactively, as shown in Figure~\ref{fig:InfiniSST-model}.

\paragraph{Streaming Speech Encoder}
We adapt the pre-trained Wav2Vec 2.0 speech encoder~\cite{baevski2020wav2vec20frameworkselfsupervised}\footnote{\url{https://dl.fbaipublicfiles.com/fairseq/wav2vec/wav2vec_vox_960h_pl.pt}} with several modifications. First, we replace the convolutional positional embedding with rotary positional embeddings (RoPE), due to its better performance on long sequence tasks. Second, we replace the original bidirectional attention with chunk-wise causal attention, where each chunk consists of 48 frames from wav2vec (equivalent to 960 ms). In chunk-wise causal attention, each frame can attend to frames within the same chunk and all preceding chunks, but not future ones. Third, to limit the context length and computational load, we use a sliding window approach, allowing chunk $i$ to attend only to the hidden states from chunks within the window $[i - w^s + 1, i]$, where $w^s=10$ represents the window size.

\paragraph{Speech-to-Token Embedding Adapter}
Outputs from the speech encoder typically have longer sequence lengths compared to the corresponding transcripts, and their embedding dimensions differ from those expected by the LLM. To address this, we incorporate two 1D convolutional layers with kernel size 2 and stride 2, effectively reducing the length of the encoder output sequence. Subsequently, a linear projection layer maps these convolutional outputs to match the LLM embedding space. This adapter downsamples input sequences by a factor of 4, converting each speech chunk of 48 frames into 12 embedding vectors.

\paragraph{Multi-turn LLM Decoder}
The decoder generates the target text and emits a special EOS token when additional speech input is needed. We utilize the Qwen2.5-7B-Instruct ~\cite{qwen2025qwen25technicalreport}\footnote{\url{https://huggingface.co/Qwen/Qwen2.5-7B-Instruct}} and structure the inputs using a multi-turn dialogue format. We also report results obtained with Llama-3.1-8B-Instruct~\cite{grattafiori2024llama3herdmodels}\footnote{\url{https://huggingface.co/meta-llama/Llama-3.1-8B-Instruct}} used as the decoder.

\subsection{Data Synthesis}

We utilize three ASR datasets for data synthesis: LibriSpeech-v12~\cite{7178964}, CommonVoice-v11.0~\cite{ardila-etal-2020-common}, and VoxPopuli~\cite{wang-etal-2021-voxpopuli}. The English transcripts are translated into Chinese and German using the 4-bit quantized \texttt{Qwen2.5-32B-Instruct} model\footnote{\url{https://huggingface.co/Qwen/Qwen2.5-32B-Instruct-AWQ}}. For LibriSpeech and VoxPopuli, whose utterances are segmented from longer speech recordings, we condition the translation on up to three preceding utterances to provide additional context. The prompt is shown below. 

\begin{verbatim}
<|im_start|>system
You are a professional translator. 
<|im_end|>
<|im_start|>user
Given an English sentence along with its
preceding sentences, translate the given 
sentence into Chinese. Do not include any 
other text.

|Preceding Sentences|
{}
|End of Preceding Sentences|

|Sentence to Translate|
{}
|End of Sentence to Translate|
<|im_end|>
<|im_start|>assistant
\end{verbatim}

Given (speech, transcript, translation) triplets, we first use Montreal Forced Aligner~\cite{mcauliffe17_interspeech} to align speech and transcript words, and then align transcript words with translation words using SimAlign~\cite{jalili-sabet-etal-2020-simalign} with LaBSE model~\cite{feng-etal-2022-language}. In this way, we obtain a mapping between speech and each translation word.

Let $m_i$ denote the right boundary timestamp of the speech segment aligned with the $i$-th translation word. To ensure monotonic alignment, we enforce $m_i = \max(m_i, m_{i-1})$. We then divide each utterance into fixed-duration chunks of 960 ms and construct a translation trajectory $(s_1, y_1), (s_2, y_2), \cdots$, where each $s_j$ is a 960 ms speech chunk and $y_j = (y_{l_j}, \cdots, y_{r_j})$ is the translation span such that $m_i \leq 960 \cdot j$ for all $i \in [l_j, r_j]$.

While segmented utterances mostly consist of clean human speech, real-world scenarios often include non-speech segments. To improve model robustness, we create robust segments by slicing unsegmented speech from LibriSpeech and VoxPopuli into 30-chunk segments\footnote{30 $\cdot$ 960 ms = 28.8 seconds.}. If a segment starts in the middle of an utterance, we shift the start to align with the utterance boundary. The trajectory for a robust segment is constructed by concatenating the trajectories of all included utterances.

Since CommonVoice consists of short, single-sentence utterances not derived from long speech, we simulate robust segments by concatenating randomly sampled utterances interleaved with randomly inserted silence intervals. 

The data statistics are shown in Table \ref{tab:data_stats}.

\begin{table}[t]
    \centering
    \begin{tabular}{c|c|c}\toprule
        Dataset & \# Robust Segments & Hours  \\\midrule
        LibriSpeech & 174112 & 1393 \\
        VoxPopuli & 85874 & 687 \\
        CommonVoice & 221717 & 1774 \\\midrule
        Total & 481703 & 3854 \\
        \bottomrule
    \end{tabular}
    \caption{Statistics of synthesized data for model training.}
    \label{tab:data_stats}
\end{table}

\begin{table*}[t]
    \centering
    \begin{tabular}{c|c|c|c}\toprule
        LLM & Data & English-Chinese & English-German \\ \midrule
        \multirow{2}{*}{Llama-3.1-8B-Instruct} & LS+CV & 39.3 / 2092 / 2691 & 21.1 / 1430 / 2183 \\
         & LS+CV+VP & 40.8 / 2159 / 2673 & 23.7 / 1503 / 2109 \\\midrule 
        Qwen2.5-7B-Instruct & LS+CV+VP & 44.3 / 2189 / 2739 & 25.1 / 1689 / 2306 \\\bottomrule
    \end{tabular}
    \caption{Translation quality and latency across different combinations of LLMs and training data evaluated on ACL60/60 development set. LS, CV, and VP refer to the LibriSpeech, CommonVoice, and VoxPopuli datasets, respectively. Metrics are reported as A / B / C, where A is BLEU, B is StreamLAAL, and C is StreamLAAL\_CA. Incorporating synthetic speech translation data from VP leads to an improvement of at least 1 BLEU point. Additionally, Qwen2.5-7B-Instruct significantly outperforms Llama-3.1-8B-Instruct in Chinese translation, with a gain of approximately 4 BLEU points.}
    \label{tab:main}
\end{table*}

\subsection{Training}
We train our model using standard cross-entropy loss on the target translation tokens, including the special EOS token, derived from robust speech segments we constructed. 
Additionally, for each robust segment, we randomly sample a latency multiplier $m \leq 12$ and merge every $m$ consecutive chunks as the data augmentation.

The training is conducted in two stages. Initially, we freeze the LLM and only train the speech encoder and adapter components. In the subsequent stage, we freeze the speech encoder, adapter and LLM, and conduct LORA finetuning~\cite{hu2022lora}.

\subsection{Inference on Unbounded Speech}
During inference, we segment the continuous input speech into fixed-length chunks of 960 ms. To manage latency, we vary the latency multiplier, ensuring translations are generated only after accumulating a predefined number of chunks. At each inference step, the newly received speech chunks are processed by both the speech encoder and the LLM, where KV caching is used to avoid redundant computations. 

The speech encoder first processes new chunks along with relevant cached context. The resulting speech features are then downsampled by the adapter into a reduced set of embeddings, matching the LLM's input requirements. The LLM subsequently generates translations based on these embeddings.

The decoder uses a sliding window strategy to maintain context, combining the cached representations of initial system instructions with the most recent generated tokens similar to ~\citet{han-etal-2024-lm}. We concatenate the KV cache of instruction with those of the most recent 1K tokens and apply RoPE on top of them. Then the LLM generates translations conditioned on this combined KV cache.

\section{Experiments}

\subsection{Setup}

We use the Adam optimizer~\cite{DBLP:journals/corr/KingmaB14} with cosine learning rate decay and 1,000 warmup steps. Training is conducted in two stages. In Stage 1, we update only the speech encoder and adapter, using a maximum learning rate of 2e-4. In Stage 2, we freeze the speech encoder and train the LLM with LoRA~\cite{hu2022lora}\footnote{rank = 32, alpha = 16, dropout = 0.1, applied to all linear layers.}, using a maximum learning rate of 1e-4. Each stage is trained for one epoch with a maximum effective batch size of 57.6K tokens. We leverage PyTorch Lightning\footnote{\url{https://github.com/Lightning-AI/pytorch-lightning}} and DeepSpeed ZeRO\footnote{\url{https://github.com/deepspeedai/DeepSpeed}} to train the model on a node of 8 NVIDIA L40S GPUs. 

During inference, we use beam search with beam size 4, repetition penalty 1.2, and ngram\_no\_repeat 5. We set test-time latency multiplier to 3 for English-to-Chinese and 2 for English-to-German. The results are evaluated with BLEU, StreamLAAL and StreamLAAL\_CA. 

\subsection{Results}

Results are presented in Table~\ref{tab:main}. While the synthetic speech translation data from LibriSpeech and CommonVoice already includes over 3K hours of speech, adding additional synthetic data from VoxPopuli consistently improves BLEU scores by at least 1 point. Moreover, replacing Llama-3.1-8B-Instruct with Qwen2.5-7B-Instruct leads to a notable gain in translation quality, particularly for English–Chinese, with an improvement of over 3 BLEU points.
\section{Conclusion}
In this paper, we presented CMU's simultaneous speech translation system built upon the InfiniSST framework for the IWSLT 2025 SST task. Our end-to-end model employs a chunkwise causal Wav2Vec 2.0 encoder, a adapter, and the Qwen2.5-7B-Instruct decoder. We demonstrated that synthesizing training data by translating large-scale ASR datasets significantly alleviates the limitations posed by limited parallel data, achieving substantial improvements in translation quality. Our experiments indicated that the addition of synthesized data from VoxPopuli provided consistent gains, and the Qwen2.5-7B-Instruct decoder notably outperformed alternatives, particularly in English-to-Chinese translation. The proposed model effectively balances translation quality and computational latency, showcasing strong performance in a realistic, unbounded speech scenario.

\bibliography{clean}

\end{document}